\documentclass[conference]{IEEEtran}
\usepackage{cite}
\usepackage{amsmath,amssymb,amsfonts}
\usepackage{graphicx}
\usepackage{textcomp}
\usepackage{enumitem}

\usepackage{bookmark}

\usepackage{pdfpages}
\usepackage{multirow}
\usepackage{lscape}

\usepackage{booktabs}
\usepackage{xcolor}
\usepackage[utf8 ]{inputenc}

\usepackage{float}
\usepackage{hyperref}
\hypersetup{ colorlinks, citecolor=green, filecolor=black, linkcolor=blue, urlcolor=blue } 
\usepackage{cleveref}
\usepackage{arydshln}
\usepackage{supertabular}
\usepackage{algorithm}
\usepackage{algpseudocode}
\usepackage{rotating}
\usepackage{listings}
\usepackage{xcolor}
\usepackage[utf8]{inputenc}

\definecolor{codegreen}{rgb}{0,0.6,0}
\definecolor{codegray}{rgb}{0.5,0.5,0.5}
\definecolor{codepurple}{rgb}{0.58,0,0.82}
\definecolor{backcolour}{rgb}{0.95,0.95,0.92}

\lstdefinestyle{mystyle}{
    backgroundcolor=\color{backcolour},   
    commentstyle=\color{codegreen},
    keywordstyle=\color{magenta},
    numberstyle=\tiny\color{codegray},
    stringstyle=\color{codepurple},
    basicstyle=\ttfamily\footnotesize,
    breakatwhitespace=false,         
    breaklines=true,                 
    captionpos=b,                    
    keepspaces=true,                 
    numbersep=5pt,                  
    showspaces=false,                
    showstringspaces=false,
    showtabs=false,                  
    tabsize=2
}

\usepackage{mdframed}

\newtheorem{definition}{Definition}[section]%

\MakeRobust{\Call}

\usepackage{xcolor}
\usepackage{authblk}
\def\BibTeX{{\rm B\kern-.05em{\sc i\kern-.025em b}\kern-.08em
    T\kern-.1667em\lower.7ex\hbox{E}\kern-.125emX}}
\begin{document}

\title {A novel post-hoc explanation comparison metric and applications}

\author[1]{Shreyan Mitra}
\author[2]{Leilani Gilpin}
\affil[1]{Paul G. Allen School of Computer Science and Engineering, University of Washington at Seattle}
\affil[2]{Department of Computer Science and Engineering, University of California at Santa Cruz}
\affil[ ]{\textit{xaisuite@gmail.com}}

\maketitle

\begin{abstract}
Explanatory systems make the behavior of machine learning models more transparent, but are often inconsistent. To quantify the differences between explanatory systems, this paper presents the Shreyan Distance, a novel metric based on the weighted difference between ranked feature importance lists produced by such systems. This paper uses the Shreyan Distance to compare two explanatory systems, SHAP and LIME, for both regression and classification learning tasks. Because we find that the average Shreyan Distance varies significantly between these two tasks, we conclude that consistency between explainers not only depends on inherent properties of the explainers themselves, but also the type of learning task. This paper further contributes the XAISuite library, which integrates the Shreyan distance algorithm into machine learning pipelines.
\end{abstract}

\section{Introduction}
From self-driving cars to customer support chat-bots, machine learning models have become pervasive in our daily lives~\cite{nirmal_2017}. The problem is that these machine learning models are opaque; the underlying model processes are not known by humans.  When these opaque systems are being entrusted with human-level decisions, e.g., sentencing convicts or driving a car, they will need to be able to explain themselves to justify their behavior~\cite{lakshmanan_2021}.

This is especially pertinent when such opaque models fail. In 2016, a ProPublica article revealed that Northpointe, a widely used criminal risk assessment tool, was racially biased. It incorrectly rated incarcerated African Americans as more likely to commit future crimes than Caucasians~\cite{angwin_larson_kirchner_mattu_2016}. And in 2018, a self-driving car hit and tragically killed a cyclist~\cite{wakabayashi2018self}. The machine learning model in the car was unable to reconcile contrasting information from various sensors, and thus failed to make the right decision~\cite{jones_2018}.

Explanatory systems produce explanations, also known as model-dependent justifications~\cite{rose2019}. They provide one way to understand machine learning models. However, for explanations to be trustworthy, it is essential that they are consistent and accurate~\cite{gilpin_2018}. Currently, there is no standard definition of explainer consistency and accuracy. Therefore, we define consistency as having two components. The first part is reproducibility - applying explanation methods repeatedly should yield the same results. Secondly, the results of different explanatory methods in similar scenarios should be similar. By accurate, we mean that the justifications provided by explanation systems are correct. In this paper, we examine two state of the art explanatory systems: SHAP~\cite{NIPS2017_7062}, based on the game theory concept of Shapley values, and LIME~\cite{lime}, which stands for local interpretable model-agnostic explanations. We propose an approach to compare explanatory systems such as SHAP and LIME and to automatically analyze cases where they are inconsistent. 

Furthermore, by highlighting the inconsistencies between explanatory systems, this paper lays the groundwork for future work that will make explanations for machine learning models more consistent and accurate. More consistent explanations will provide users and stakeholders a supported reason behind system malfunctions, preventing incidents like the one involving the self-driving car and the cyclist. In addition, system debugging and diagnosis will be more efficient. 

Consistent explanations are the first step towards trustworthy explanations. Without trustworthy explanations, users are effectively blind to the operation of machine learning models and cannot mitigate model flaws. Our work therefore answers the following research questions: 

\begin{enumerate}[leftmargin=*]
	\item How can consistency between explanatory systems be measured?  
	\item How can the metric calculated in (1) be used to draw generalized conclusions about the consistency of explanatory systems for a given learning task?
    \item How can comparison of explanatory systems be integrated into conventional machine learning pipelines?
\end{enumerate}

In this paper, we present the Shreyan Distance and the XAISuite Library, both of which attempt to answer these questions. For example, the Shreyan Distance measures consistency between explanatory systems by comparing lists of features ranked by importance values by different explainers for a particular instance. This is a potential solution to Question 1. In addition, by using the Shreyan Distance to identify patterns in explainer similarities across 64 models and 2 different learning tasks, we seek to answer Question 2. And finally, the XAISuite library offers a seamless way to integrate the Shreyan Distance with other common machine learning utilities through wide-ranging compatibility and a comprehensive model-selection-to-explanation framework. This is our response to Question 3. 

\section{Background/Related Work}
 Our paper builds on previous work in vector comparison, explanations, failure analysis, and machine learning error.

\subsection{Vector Comparison} 
In this paper, we compare vectors containing features ordered by importance. Several methods to compare vectors exist. 

The Spearman's Distance is the square of the Euclidean distance bteween two ranked vectors, and is often used as a measure of disarray \cite{10.1111/j.2517-6161.1977.tb01624.x}. However, it cannot be easily applied to comparing ordered feature importance lists produced by explainers. This is because Spearman's Distance does not weight differences in certain dimensions of the vectors more than differences in others. However, in explanation comparison, differences in what the explainers detect as the most important feature are more significant than differences for other features. The Shreyan Distance rectifies this by assigning a linear weightage to each vector dimension.

Kendall's rank correlation coefficient (also referred to as Kendall's Tau) poses the same problem as Spearman's Distance, i.e. it fails to take into account weight differences between different vector dimensions. While it is less sensitive to variations in data, it is affected equally by equal differences in different vector dimensions. A weighted Kendall's Tau statistic \cite{SHIEH199817} has been proposed. However, there are important differences between a weighted Kendall's Tau and Shreyan's Distance. The lowest value of the Kendall's Tau metric is -1, which indicates a perfect negative relationship between two variables. However, the lowest value of the Shreyan Distance, 0, does not necessarily indicate a perfect negative relationship, but the highest cumulative weighted difference. The authors believe that, in the context of explainers, the Shreyan Distance is a better metric of the difference because of user expectations, as elaborated in the Results Section. An example of such an user expectation would be to measure SHAP and LIME as having similarity if they identify the same feature for a given importance rank, even if their feature importance lists are negatively correlated overall.






\subsection{Metrics for Explanation Accuracy} 

 Recall from the introduction that explainer accuracy is how similar the justifications produced by explanation systems are to reality. In this section, we provide an overview of existing explanatory accuracy baselines.

 A research paper published by DeepMind \cite{https://doi.org/10.48550/arxiv.1706.08606} suggests that suggests that machine learning is an analog to human thought and can be explained through similar processes. The paper explores the use of cognitive psychology to explain the decisions of machine learning models, drawing parallels between biases humans develop during their maturation and those acquired by machine learning. By likening machine learning models to humans, the paper provides a framework to determine which explanations have a higher probability of being accurate. Since explanations are ultimately meant for human understanding, we find the use of psychology in explanation generation promising and perhaps capable of resolving the discrepancies between explanatory systems found in this paper. 

 Gilpin et al. \cite{gilpin_2018} believe that what is defined as accurate might depend on user requirements. They note a tradeoff between completeness and interpretability that all explanatory systems must follow - the more accurate explanations are, the less likely they are to be understandable by humans. This tradeoff may affect the discrepancies between different explanatory models. Thus, a key part of future explanatory system research is creating explainers that gain the user's trust~\cite{goel_sindhgatta_kalra_goel_mutreja_2022}. We see our work in quantifying explainer consistency as integral to that effort. Research on user requirements for explanations is elaborated on further later in this section.

 Han et al. \cite{han2022} propose that different explanatory systems are optimal for different scenarios, and an ``adversarial" sample exists that will lead to a large error for any given explanatory system. For example, while SHAP and LIME are both based on local function approximation (LFA), they differ in their optimal intervals due to their noise functions. This can help explain the discrepancies between SHAP and LIME that we observe in our research for different learning tasks. Research on explanation comparison is explored later in this section.

\subsection{On user requirements for meaningful explanations}

 Numerous papers \cite{riveiro2021,10.1145/1943403.1943424,10.1145/3503252.3531306} have highlighted the importance of user expectations in explanation utility. Since user expectations are often implicit, determining what type of explanations users are looking for is difficult. The XAISuite library proposed in this paper alleviates this problem by providing users with the option to use multiple explanatory systems, compare them, and choose the explanations most suitable to their scenario. 

 Chazette et al. \cite{9920064} previously created a framework for explanatory systems embedded in user trust. Consistency was a key factor in their analysis, and the XAISuite library is designed keeping the requirements outlined in the paper in mind.  

 Machine learning models need to be safe and
    trustworthy, especially when entrusted with human-level decision
    making. \cite{10.1007/978-3-642-32378-2_8}. One way to ensure safety is to have stricter requirements and guarantees. In 2021, Nadia Burkart and Marco Huber \cite{burkart_2021} laid out the requirements of explainable supervised machine learning models. Our paper implements two of their requirements: (1) We make explanation of machine learning models more easily available through our open source XAISuite library and (2) By highlighting inconsistencies between explanatory systems, we set the scene for more consistent and trustworthy explainers. 

  User requirements for explanations may vary in specialized fields. Ghassemi et al. \cite{Ghassemi2021} argue against the use of explainers in the medical profession, claiming that the many failures and contradictions of explanatory systems endanger the trust of healthcare professionals and the lives of patients. They propose that machine learning models be rigorously tested instead. However, we believe that the solution to explainer error is not abandoning explanatory systems altogether, but to improve them until they are trustworthy. Our analysis of explainer consistency is a step in this direction because it highlights contradictions among different explanatory systems and identifies particular test cases that lead to explainer inconsistency.  

\subsection{Explaining machine learning failure} 

 Examples of machine learning failure abound, but Gilpin et.al \cite{gilpin_2018} specifically note several cases where machine learning systems fail, including racially biased criminal-assessment tools and flawed categorization due to the introduction of noise. For users to trust ML systems, they need to be able to understand the rationale behind mistakes made by machine learning models on such cases where the input leads to machine learning errors~\cite{https://doi.org/10.48550/arxiv.1312.6199}. A user of the XAISuite library has the ability to consult different explainers, a key part of gaining more insight about such ``adversarial" cases. 

\subsection{On comparing explanatory systems}

 Comparing explanatory models is an open area of research. This is sometimes known as the ``disagreement problem" \cite{https://doi.org/10.48550/arxiv.2202.01602}. 

 In their paper, Covert et al. \cite{covert2022} point out that while there are many different explanatory methods, it remains unknown how ``most methods are related or when one method is preferable to another." The authors propose a new class of similar explanations supported by cognitive psychology called removal-based explanations. These systems determine the importance of a feature by analyzing the impact of its removal. The paper specifically highlights that as SHAP and LIME are both part of the removal-based explanatory framework, they share a resemblance. The discrepancies between SHAP and LIME shown in our paper are markers of where two very similar explanatory systems with related internal mechanisms can differ. 

 Roy et al. \cite{9978217} in ``Why Don't XAI Techniques Agree?" acknowledged that SHAP and LIME explanations often disagree and that users do not know which method to trust. They proposed an aggregate explainer that focuses on the similarities between SHAP and LIME and disregarded discrepancies. But the authors of that paper do not set forth a way to find and resolve the discrepancies. We contribute a
 method to empower users to better identify points of disagreement between SHAP and LIME and come to a decision. 

 Van der Waa et al. \cite{VANDERWAA2021103404} extended explanatory system comparison further with a detailed analysis of rule-based versus example-based explanations, with implications on user trust and accuracy. Our contribution allows users to validate this analysis by highlighting the differences between explanatory systems rather than trusting only one of them. 

 Duell et al. \cite{9508618} specifically compares the results of explanatory systems such as SHAP and LIME on electronic health records. They note significant differences in importance scores between the explanatory systems, stating that ``studied XAI methods circumstantially generate different top features; their aberrations in shared feature importance merit further exploration from domain-experts to evaluate human trust towards XAI." While this aligns with the results of our paper, we extend the study to different types of data outside of health records. We also create a generalized machine learning pipeline to help in the ``further exploration'' Duell et al. deemed necessary. 

 The in-depth comparison that we perform between two explanatory systems has been previously explored. A paper published by Lee et al. \cite{Lee2022} compares breakDown (BD) and SHAP explainers in the specific case of classification of multi-principal element alloys. However, their work is not generalizable to all tabular data and all machine learning models, which our contribution is. That paper also does not propose a systematic way to compare explainers, which we do. In our work, we use various models on data of different types to allow us to have a better picture of exactly what factors affect explainer consistency. Furthermore, we compare SHAP and LIME, two of the most commonly used machine learning models in the field, and thus our results are more applicable to different machine learning tasks. 

\subsection{Related Software}

 Various tools similar to XAISuite also exist. However, the contribution of a novel explanation comparison metric is unique to XAISuite. 

 \begin{enumerate}[leftmargin=*]
 	\item{Agarwal et al. \cite{agarwal2022openxai} created a tool, OpenXAI, for evaluating and benchmarking post-hoc explanation systems, comparable in functionality and user interface to our XAISuite. While OpenXAI focuses more on accuracy of explanatory systems over one another in specific tasks, we put a heavier emphasis on consistency for all tasks. Furthermore, we integrate explanatory comparison into XAISuite. }
	
 	\item{Yang et al. \cite{wenzhuo2022-omnixai} created the OmniXAI library for explainable AI that allows easy access to numerous explainers for a particular machine learning model. The OmniXAI library serves as part of the backend of the XAISuite library by helping to fetch explanatory systems that the user requests. The XAISuite library's data visualization and explanation comparison abilities build on the core functionality offered by OmniXAI.}
	
 	\item{Captum, similar to OmniXAI, was proposed by Kohklikyan et al. \cite{2009.07896}. In their implementation of the XAISuite library, the authors believed that OmniXAI was easier to work with.}
 \end{enumerate}

\section{Methods}\label{sec11}

We take a theoretical, experimental, and a software development approach in this paper. Therefore, we divide the methodology into three parts: (1) developing the Shreyan Distance metric (2) using the metric to compare SHAP and LIME values for classification and regression, and (3) building an adaptable software system: the XAISuite Framework.







\subsection{The Shreyan Distance}

Here, we define a metric called the Shreyan Distance for comparing the results produced by explanatory systems. 

Let the quantity $d_{max}$ for two ordered vectors each of equal size $x$ be defined as:
\newline
\begin{align*}
d_{max}\left(x\right)&=\ \sum_{n=1}^{\operatorname{floor}\left(\frac{x}{2}\right)}\frac{\left(\left(x-n\ +\ 1\right)\cdot\left(x-2n+1\right)\right)}{x^{2}}\ +\ \\
&\sum_{n=\operatorname{floor}\left(\frac{x}{2}\right)+1}^{x}\frac{\left(x-n+1\right)\cdot\operatorname{floor}\left(\frac{x}{2}\right)}{x^{2}}
\end{align*}

Let the two vectors be $r$ and $r\ast$. Then, we define the quantity $d$ for the two vectors as:
\newline
\newline

\begin{align*}
d\left(r,\ r\ast\right)\ &=\ \sum_{n=1}^{x}\left(x-n+1\right)\cdot\frac{\left|r\left(n\right)\ -\ r\ast\left(n\right)\right|}{x^{2}}
\end{align*}
\newline
\newline
The Shreyan Distance is defined below:
\newline
\begin{definition}[Shreyan Distance]
\newline
\newline
$d_{s}\left(r,\ r\ast\right)\ =\ 1 - \frac{d\left(r,\ r\ast\right)}{d_{max}\left(\operatorname{length}\left(r\right)\right)}$
\newline
\end{definition}
The Shreyan Distance has desired properties, which we illustrate in the following examples. We also compare the Shreyan Distance with  Spearman's Distance and weighted Kendall's Tau in some cases.
\newline
\subsubsection{Example 1: A Simple Shreyan Distance Calculation}
Take the following feature importance lists produced by two explainers, Explainer 1 and Explainer 2. The lists have already been enumerated for simplicity.
\newline
\newline
Explainer 1: [1, 2, 3, 4, 5]
\newline
Explainer 2: [2, 1, 3, 5, 4]
\newline
\newline
Here, $r$ = [1, 2, 3, 4, 5], $r\ast$ = [2, 1, 3, 4, 5], and $x$ = length($r$) = length($r*$) = 5
\newline
\newline
For this vector pair, 
$d_{max}$ = 1.6, $d$ = 0.48, and $d_s$ = $1 - 0.48/1.6$ = 0.7
\newline
\newline
We take a Shreyan Distance value of 0.7 to indicate moderate similarity. This example demonstrates the ability of the Shreyan Distance algorithm to generate an illustrative number which can be used to draw conclusions about the similarity between two explainers. 
\newline
\subsubsection{Example 2: Comparison with Spearman's Distance}
Here are  three vectors:
\newline
\newline
Vector 1: [1, 2, 3, 4, 5]
\newline
Vector 2: [1, 2, 3, 5, 4]
\newline
Vector 3: [2, 1, 3, 4, 5]
\newline 
\newline

The Spearman's Distance between Vector 1 and Vector 2 and the Spearman's Distance between Vector 1 and Vector 3 are the same (the value is 2 for both). But this is not realistic in an explainer importance context. We would expect Vector 1 and Vector 2 to be more similar. This is because users care more about consistency in the most important feature detected by different explainers than consistency is less important features. The vector elements are ordered by descending importance, so Vector 1 and Vector 2's differences are smaller than those between Vector 1 and Vector 3. This is reflected by the Shreyan's Distance:
\newline
\newline
Between Vector 1 and Vector 2: $d_s$ = 0.925
\newline
Between Vector 1 and Vector 3: $d_s$ = 0.775
\newline
This example showcases how Shreyan's Distance is catered specifically to the comparison of explanatory features and other ranked vectors, as compared to Spearman's distance. 
\subsubsection{Example 3: Negative Correlation Does Not Result in Minimum Shreyan Distance}
Take the following feature importance lists produced by two explainers, Explainer 1 and Explainer 2. 
\newline
\newline
Explainer 1: [1, 2, 3, 4, 5]
\newline
Explainer 2: [5, 4, 3, 2, 1]
\newline
\newline
The two vectors are perfectly negatively correlated. They have a Kendall's Tau value of -1. 
\newline
We might expect the Shreyan Distance to similarly be the lowest, i.e. 0. But this is not the case. $d_s$ for these two vectors is 0.1. The authors believe that when users look to compare explanations, they are looking for whether explainers choose the same features as being important, not about general trends of the order of the features. The 0.1 value of the Shreyan Distance is therefore an acknowledgement of the fact that both Explainers 1 and 2 share element 3 as the third most important feature. 

\subsubsection{Example 4: Shreyan Distance Distribution and Pearson Correlation Distribution}
Below, we include a graph of a 1000-sized sampling distribution of values for the Shreyan Distance and Pearson Correlation with $x$ = 9. The Pearson Correlation is normalized to fit within the bounds 0 to 1. While the distribution of Pearson Correlation Coefficients is roughly normal, the Shreyan Distance Distribution has a right skew and a center less than 0.5. This indicates that, compared to the Pearson Correlation Coefficient, the Shreyan Distance is more likely to rate two explainers as being different. When it comes to explainer consistency, the authors believe that it is better to err on the side of caution. This is reflected by the higher probability of the Shreyan Distance being less than 0.5 over being greater than 0.5. 

\begin{figure}[!htbp]
    \vspace{-2ex}
    \centering
    \includegraphics[scale=0.4]{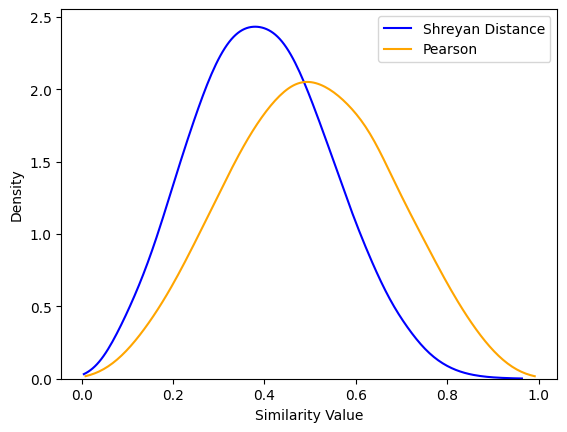}
	\caption{"Density Plots for Shreyan Distance and Pearson, given $x$ = 9"}
\end{figure}

\subsection{Comparing SHAP and LIME for Regression and Classification Tasks}
The models we use in our analysis, along with the task for which they will be used, are listed in \hyperref[tab]{Table 1}. These models were chosen to represent a wide variety of machine learning models so as to prevent any confounding variables.

\begin{table*}[]
\label{tab}
\resizebox{\textwidth}{!}{%
\begin{tabular}{@{}cl@{}}
\toprule
Task &
  \multicolumn{1}{c}{Models} \\ \midrule
Regression &
  \begin{tabular}[c]{@{}l@{}}ARDRegression, AdaBoostRegressor, BaggingRegressor, BayesianRidge, DecisionTreeRegressor, DummyRegressor, \\ ElasticNet, ElasticNetCV, ExtraTreeRegressor, ExtraTreesRegressor, GaussianProcessRegressor, GradientBoostingRegressor, \\ HistGradientBoostingRegressor, HuberRegressor, KNeighborsRegressor, KernelRidge, Lars, LarsCV, Lasso, LassoCV, LassoLars, \\ LassoLarsCV, LinearSVR, MLPRegressor, NuSVR, OrthogonalMatchingPursuit, OrthogonalMatchingPursuitCV, PLSRegression, \\ PassiveAggressiveRegressor, PoissonRegressor, QuantileRegressor, RandomForestRegressor, Ridge, RidgeCV, SGDRegressor, \\ SVR, TheilSenRegressor, TransformedTargetRegressor, TweedieRegressor\end{tabular} \\ \midrule
Classification &
  \begin{tabular}[c]{@{}l@{}}AdaBoostClassifier, BaggingClassifier, BernoulliNB, CalibratedClassifierCV, DecisionTreeClassifier, DummyClassifier, \\ ExtraTreeClassifier, ExtraTreesClassifier, GaussianNB, GaussianProcessClassifier, GradientBoostingClassifier, \\ HistGradientBoostingClassifier, KNeighborsClassifier, LinearDiscriminantAnalysis, LinearSVC, LogisticRegressionCV, \\ MLPClassifier, NuSVC, Perceptron, QuadraticDiscriminantAnalysis, RandomForestClassifier, RidgeClassifier, RidgeClassifierCV, \\ SGDClassifier, SVC\end{tabular} \\ \bottomrule
\end{tabular}%
}
\caption{Machine learning models used for regression and classification tasks. All models were used with default, non-optimized parameters.}
\label{tab:my-table}
\end{table*}

To compare SHAP and LIME for 64 models across both regression and classification tasks, we follow the procedure below: 

\begin{enumerate}
\item Install the XAISuite library (to be discussed in Section C) from the Python Package Index.
\item Install the SciKit-Learn library from the Python Package Index if it is not already automatically installed through XAISuite. 
\item Install the SciPy library from the Python Package Index. 
\item Let $regData$ be an array holding data from the regression tasks and $classData$ be an array holding data from the classification tasks.
\item Regression Tasks: For each regression model in \hyperref[tab]{Table 1},  
\begin{itemize}
\item Use the XAISuite library to calculate the average Shreyan Distance scores on a regression dataset of size $n$ generated by SciKit-Learn. In our research, we take $n$ = 100.  Regenerate the dataset for each iteration. 
\item Repeat this three times for each model. Append all three Shreyan Distance results to $regData$
\end{itemize}
\item Classification Tasks: For each classification model in \hyperref[tab]{Table 1},
\begin{itemize}
    \item Use the XAISuite library to calculate the average Shreyan Distance scores on a classification dataset of size $n$ generated by SciKit-Learn. In our research, we take $n$ = 100. Regenerate the dataset for each iteration. 
    \item Repeat this three times for each model. Append all three Shreyan Distance results to $classData$
\end{itemize}
\item Use the stats package of SciPy to perform a 2-sample t-test for difference in population means using $regData$ and $classData$ as the two samples. We assume that the scores in $regData$ and $classData$ are independent and have equal variance. Use a $\alpha$ = 0.05 or 5\% significance level. 

For the t-test, use the following hypotheses:
\newline
Null Hypothesis $H_o$: There is no difference between the true mean LIME and SHAP similarity (as measured using the Shreyan Distance) between classification and regression tasks. 
\newline
Alternate Hypothesis $H_a$: There is a difference between the true mean LIME and SHAP similarity (as measured using the Shreyan Distance) between classification and regression tasks. 
\item Interpret the results of the statistical test to determine whether there is significant evidence that the true SHAP and LIME similarity, as measured by the Shreyan Distance, is different for regression and classification tasks.
\end{enumerate}

Below, we include setup code that implements the procedure above:

\begin{lstlisting}[language=Python, caption=Setup script for SHAP and LIME comparison]
from xaisuite import*
import seaborn as sns
import scipy.stats as stats
#The variable definitions for reg_models and class_models are not included here for the sake of space. See Table 1. 
regData = []
classData = []
for model in reg_models:
  for i in range(3):
      corr = InsightGenerator(ModelTrainer(model, DataProcessor(DataLoader(make_regression) , processor = "TabularTransform"), explainers = {"lime": {"feature_selection": "none"}, "shap":{}}).getExplanationsFor([])).calculateExplainerSimilarity("lime", "shap")
      regData.append(corr)   
for model in class_models:
  for i in range(3):
      corr = InsightGenerator(ModelTrainer(model, DataProcessor(DataLoader(make_classification) , processor = "TabularTransform"), explainers = {"lime": {"feature_selection": "none"}, "shap":{}}).getExplanationsFor([])).calculateExplainerSimilarity("lime", "shap")
      classData.append(corr)
sns.kdeplot(regData, label = 'Regression Distribution')
sns.kdeplot(classData, label = 'Classification Distribution')
stats.ttest_ind(a=regData, b=classData, equal_var=True)
\end{lstlisting}
\subsection{The XAISuite Library and Framework}
We now present the XAISuite framework, a unified tool for comparing explanatory systems. It forms the basis of the XAISuite library, which enables users to train and explain models with minimal input. It also answers one of our primary research questions by integrating the Shreyan Distance metric into machine learning pipelines. In constructing this library, we build on OmniXAI \cite{wenzhuo2022-omnixai}, which allows direct access to different explainers. 

\begin{figure}[!htbp]
    \label{framework}
    \vspace{-2ex}
    \centering
    \includegraphics[scale=0.4]{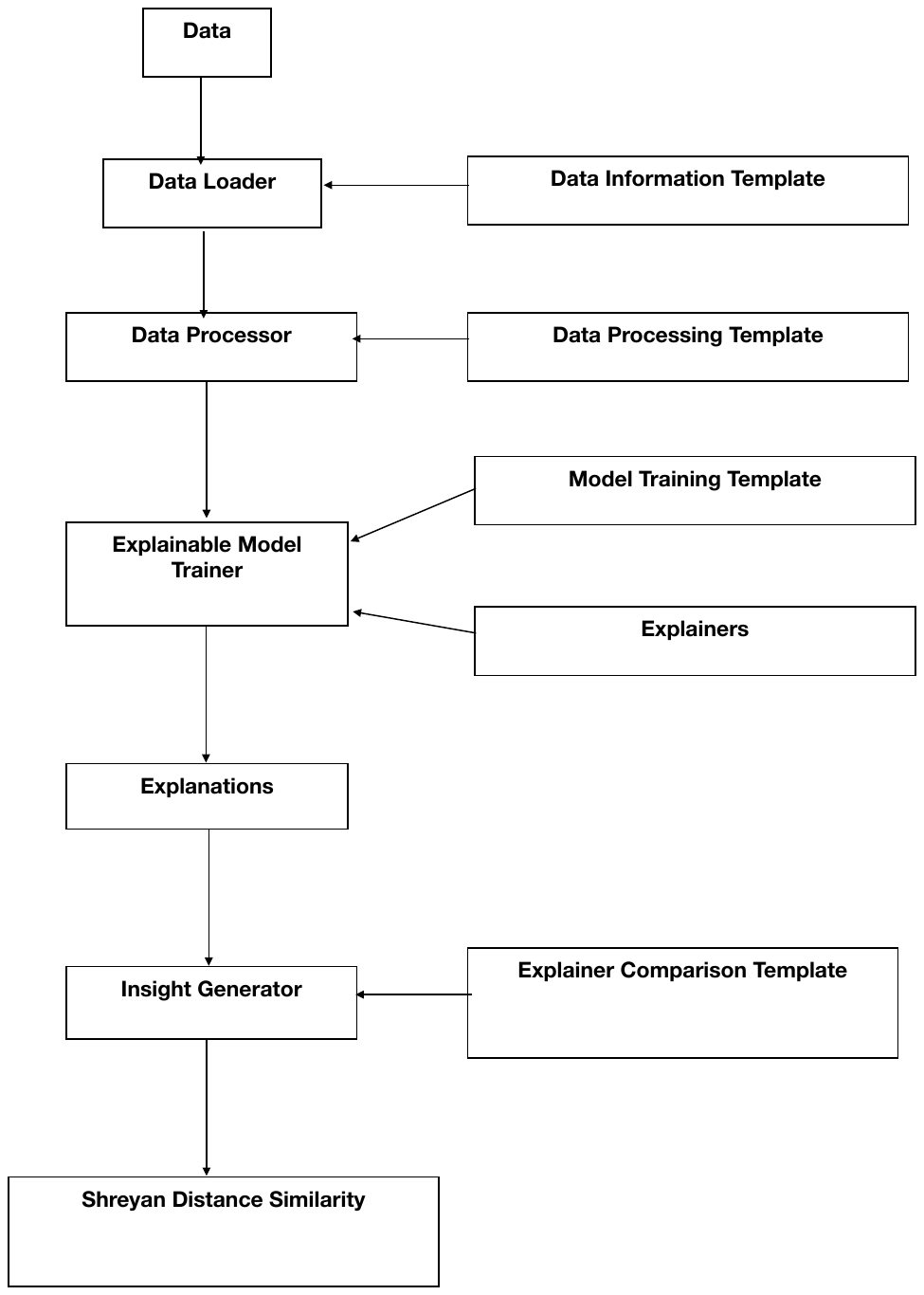}
	\caption{The XAISuite Framework: Integrating the Shreyan Distance in to a comprehensive machine learning pipeline. The library can be found at \href{github.com/11301858/XAISuite}{github.com/11301858/XAISuite}}
\end{figure}

A brief overview of the XAISuite framework is presented in \hyperref[framework]{Figure 2}. The framework consists of four components: data loading and processing, machine learning model training and explanation, and explanation generation.

We contribute the XAISuite framework with the intention for it to be a standard platform for training, explaining, and analyzing models. The framework was constructed based on five key factors:

\begin{enumerate}[leftmargin=*]
	\item \textbf{Simplicity: } Containing just three parts which depend on data retrieval, function calls, and writing to output files, XAISuite provides guiding principles for any implementation to enhance code changes and user convenience.
	\item \textbf{Integratablity: } A library is limited if it cannot be used with other libraries. Core functionalities, like model training, explanation generation, and graphics creation, are designed to use external libraries. However, the framework is flexible and not based on any specific external dependency, so there is flexibility in the way in which the model is trained or explanations are generated.
	\item \textbf{Flexibility: } A key feature of XAISuite is flexibility. This is enabled by the lack of specifics and the use of general terms. Note that \emph{any} dataset can be used, \emph{any} model can be trained, \emph{any} explanatory system can be initialized, depending on the implementation. The templates have no fixed form, nor is the form of data storage specified. As mentioned in the previous point, the XAISuite framework is compatible with any potential provider of its constituent parts, whether they be model libraries(sk-learn, XGBoost, etc.), transform function types (Logarithmic, Exponential, etc.), or different data storage options (Dataframe, Numpy, Files, etc.).
	\item \textbf{Usability: } Users are the center of explainability research \cite{riveiro2021}, and so any interface that facilitates interactions between the user and explanatory system must be user-centric. XAISuite achieves this by ensuring that results are understandable in a readable table or graphical format. Again, individual implementations of data or graph generation may vary, but by enforcing the requirement of converting data into a portable and visualizable medium, XAISuite reinforces the human-centric approach that is a hallmark of explainability research.
	\item \textbf{Expandability: } XAISuite is not designed to be a closed system. There are opportunities for users to extend existing functionalities or link XAISuite with other existing frameworks.
\end{enumerate}

Algorithms and code for the implementation of the XAISuite Framework's machine learning model training and explanation utilities is included in the XAISuite codebase.

\section{Results}

The Shreyan Distance is a novel way to quantify the differences between the results of explanatory systems. We used the average Shreyan distance to evaluate explainer similarity across several models and datasets for regression and classification tasks. Shown below is a bar graph representing the calculated explainer similarities for each instance of an autogenerated dataset used to train an AdaBoostClassifier model. The calculated average is also shown as a red line.

\begin{figure}[!htbp]
    \label{fig2.5}
    \vspace{-2ex}
    \centering
    \includegraphics[scale=0.4]{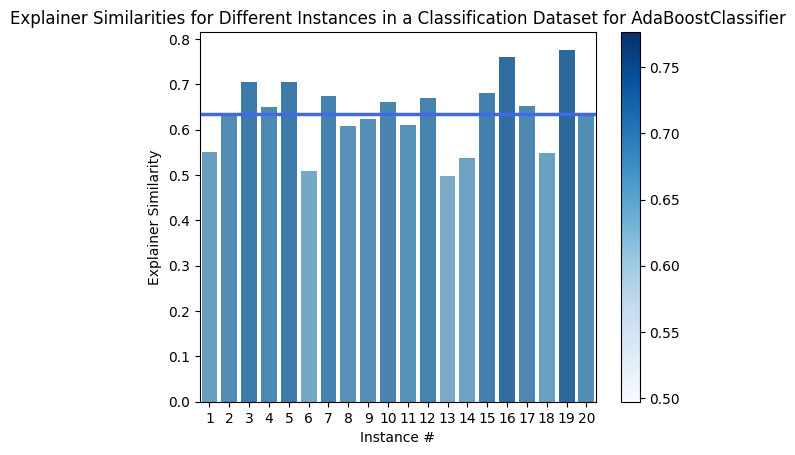}
	\caption{The model was trained on a classification dataset with 20 features and 2 classes. }
\end{figure}

We then conducted a 2-sample t-test for difference between population means to determine whether there is a statiscally significant difference between the average LIME-SHAP similarity (as measured by the Shreyan Distance) for regression and classification tasks. 
The test yielded a test statistic of -2.58 and a p-value of 0.01. Because 0.01 $<$ 0.05, we can reject the null hypothesis. There is convincing evidence that there is a difference between the true mean LIME and SHAP similarity (as measured using the Shreyan Distance) between classification and regression tasks.

\begin{figure}[!htbp]
    \label{fig3}
    \vspace{-2ex}
    \centering
    \includegraphics[scale=0.4]{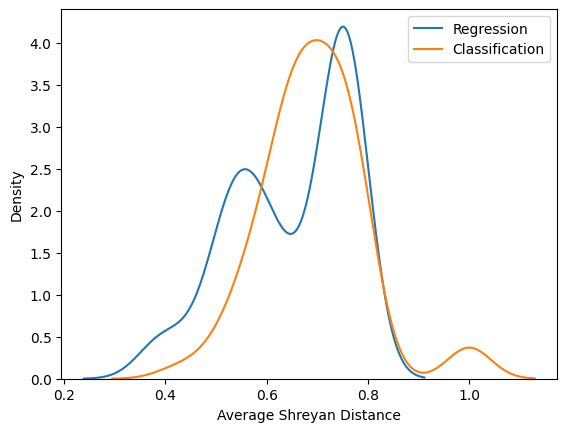}
	\caption{Distributions of Shreyan Distance for regression and classification tasks}
\end{figure}

 \hyperref[fig3]{Figure 3} is a density plot showing the distribution of the Shreyan Distance values for the regression and classification models. 

The data in \hyperref[resultstab]{Table 2} contains summary statistics about the average Shreyan Distances for the regression and classification tasks. 
\begin{table*}
\caption{Summary data from the study}
\label{resultstab}
\centering
\begin{tabular}{lll}

\toprule
{} &                                 Regression &             Classification \\
\midrule
minmax   &  (0.37273575413884535, 0.7773132717460898) &  (0.4242825607064017, 1.0) \\
mean     &                                   0.649809 &                   0.692116 \\
variance &                                   0.013377 &                   0.010448 \\
skewness &                                  -0.590239 &                   0.560591 \\
kurtosis &                                  -0.839749 &                    1.85009 \\
\bottomrule
\end{tabular}
\end{table*}

The Shreyan Distances were calculated by the XAISuite library, which we developed in the Methodology section. This supports the claim that explainer comparison can be integrated into machine learning pipelines. 

\section{Discussion}
The results showed that similarity between SHAP and LIME importance scores differ significantly for regression and classification tasks. While previous studies have shown that explainers are inconsistent, our work shows that the level of inconsistency is different for different types of learning tasks. Therefore, any explainer inconsistency is not just an inherent property of the explainers themselves, but it depends also on the training data and type of model being explained. 

The Shreyan Distance by itself also represents an advancement in the field of machine learning explainability. By taking into account user-expectations and offering advantages over other metrics like Spearman's Footrule or Kendall's Tau to measure explainer similarity, the Shreyan Distance provides a way to accurately differentiate varying accounts of what the ground truth is. We lay the groundwork for progress towards greater explainer consistency by quantifying these explainer inconsistencies through our experimental results and by proposing Shreyan's Distance, 

In addition to the XAISuite Framework and the results of our study, we would also like to briefly mention several related projects that we believe have relevance to a discussion about the XAISuite library. XAISuiteCLI is a comprehensive machine learning explainability command-line tool keeping with the XAISuite framework's emphasis on usability. This utility allows users to train and explain machine learning models using shell commands.  XAISuiteGUI is a comprehensive graphical user interface that allows users without coding experience to train, explain, and compare machine learning models. XAISuiteBlock is a block-based site inspired by Scratch for machine learning model training and explanation. This offers machine learning utilities to those without coding experience. This is a great step forward in making machine learning explainability available to everyone regardless of age or coding experience. We envision it as a potential educational tool. 

\section{Conclusion}
Explanatory systems allow users to look through the ``black box" of machine learning models. This is not only useful in understanding the internal mechanisms of machine learning models but also is essential in diagnosing model malfunctions. However, when multiple explanatory systems differ, users do not know which one to trust.
By arming users of machine learning with the information they need to make decisions, XAISuite increases trust. Furthermore, through its contribution of several interfaces catered to people regardless of age or coding experience, XAISuite empowers the use of machine learning among those that would be previously unable to do so. Finally, by setting the example for a comprehensive framework on machine learning explainability, XAISuite makes the entire process of machine learning more transparent and understandable.

Our work opens up new areas of possible research in this regard. While we performed our analysis with  two types of tasks and 64 models, we encourage others to replicate our work with more models, more learning tasks, larger datasets, and more explainers to confirm our results. With regards to the Shreyan Distance, we look forward to future work evaluating it in comparison with other existing comparison algorithms and exploring new applications. In addition, we understand that SHAP and LIME are inherently mathematical models, and we look forward to a mathematical basis for the results of our study. 
Our goal is that the results of this paper, along with the XAISuite framework and algorithm we outline, will facilitate further efforts to resolve discrepancies between explanatory systems so that humans can gain a clearer understanding of how machine learning models work. This will lead, in turn, to a greater ability to fine tune these models to prevent error.

\noindent

	
	

\bibliographystyle{IEEEtran}
\bibliography{IEEEabrv,references}

\end{document}